\pdfoutput=1

\documentclass[11pt]{article}

\usepackage[]{acl}

\usepackage{times}
\usepackage{latexsym}

\usepackage[T1]{fontenc}

\usepackage[utf8]{inputenc}

\usepackage{microtype}

\usepackage{algorithm}
\usepackage{algorithmic}

\usepackage{graphicx}
\usepackage{booktabs}
\usepackage{multirow}
\usepackage{subcaption}
\usepackage{amsmath}
\usepackage{amssymb}

\title{Patient Outcome and Zero-shot Diagnosis Prediction with \\ Hypernetwork-guided Multitask Learning}

\author{Shaoxiong Ji and Pekka Marttinen \\
Aalto University, Finland \\
  \texttt{\{shaoxiong.ji;~pekka.marttinen\}@aalto.fi} \\
  }

\begin{document}
\maketitle
\begin{abstract}
Multitask deep learning has been applied to patient outcome prediction from text, taking clinical notes as input and training deep neural networks with a joint loss function of multiple tasks.
However, the joint training scheme of multitask learning suffers from inter-task interference, and diagnosis prediction among the multiple tasks has the generalizability issue due to rare diseases or unseen diagnoses.
To solve these challenges, we propose a hypernetwork-based approach that generates task-conditioned parameters and coefficients of multitask prediction heads to learn task-specific prediction and balance the multitask learning.
We also incorporate semantic task information to improve the generalizability of our task-conditioned multitask model. 
Experiments on early and discharge notes extracted from the real-world MIMIC database show our method can achieve better performance on multitask patient outcome prediction than strong baselines in most cases.
Besides, our method can effectively handle the scenario with limited information and improve zero-shot prediction on unseen diagnosis categories.
\end{abstract}

\section{Introduction}

Recent advances apply artificial intelligence to predict clinical events or infer the probable diagnosis for clinical decision support~\cite{shickel2017deep,li2021neural}. 
Those works extensively study clinical data from Electronic Health Records (EHRs). 
For example, the Doctor AI model for predictive modeling on EHRs~\cite{choi2016doctor} and 
the Deep Patient model for unsupervised patient record representation learning~\cite{miotto2016deep} is proposed by leveraging longitudinal medical data. 
Within the patient records, we can also find various clinical notes written by general practitioners or physicians and annotated with specific diagnosis results, medical codes, and other clues of patient outcome.
Those clinical notes consist of meaningful health information, including health profile, clinical synopsis, diagnostic investigations, and medications, which can empower the clinical decision making~\cite{li2021neural}. 
Clinical text mining has demonstrated its feasibility in diverse healthcare applications, including medical code prediction~\cite{ji2021medical}, readmission prediction~\cite{huang2019clinicalbert} and diagnosis prediction~\cite{van2021clinical}. 
Readmission and length-of-stay prediction can help with utility management, especially when healthcare service is in high demand. 
Diagnostic prediction can support clinicians in making decisions based on many clinical reports. 
This study focuses on multitask patient outcome prediction from clinical notes, which classifies patients' diagnostic results and predicts clinical outcomes to empower expert clinicians and improve healthcare services' efficiency.

\begin{figure}[htbp!]
\centering
	\includegraphics[width=0.35\textwidth]{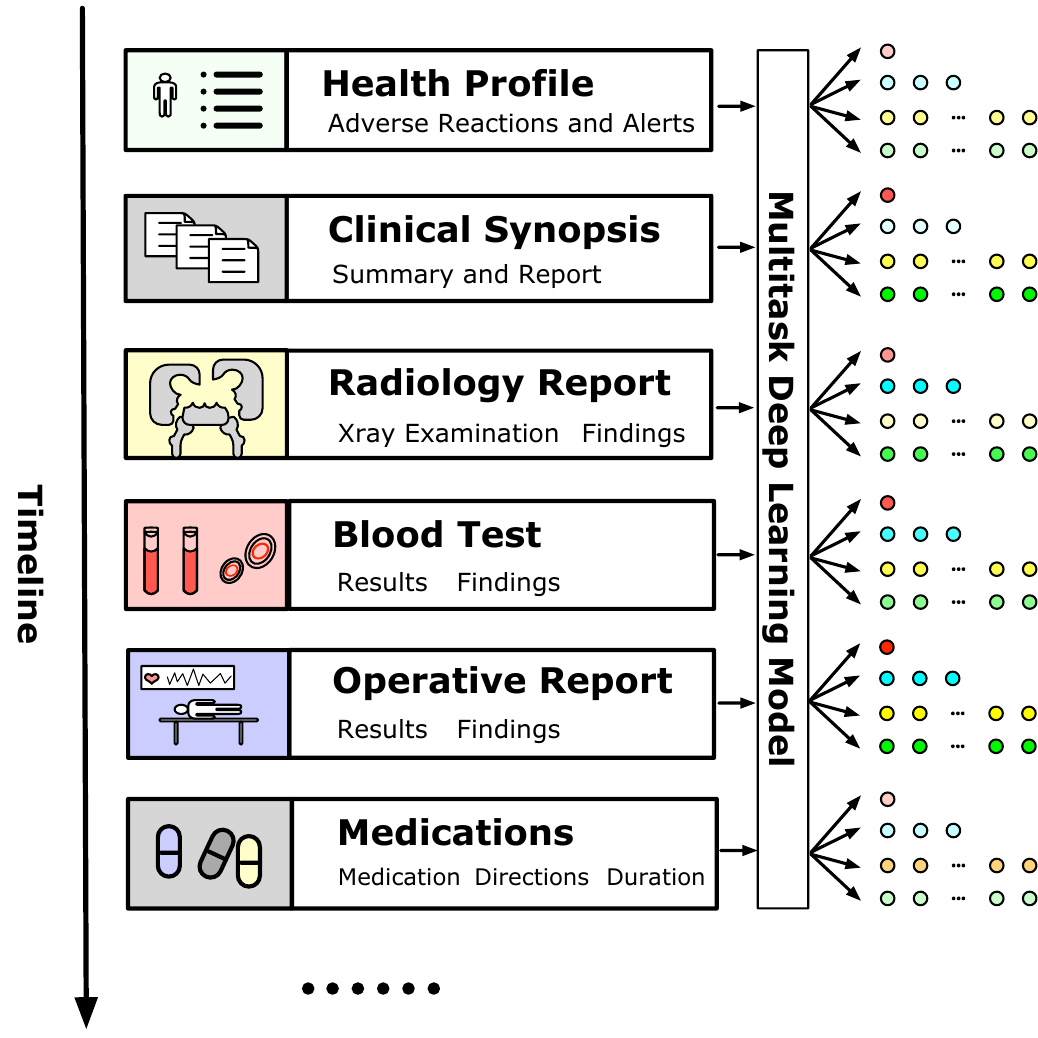}
	\caption{An example of multitask patient outcome prediction based on sequential inputs of clinical notes in the electronic health record. The multitask deep learning model dynamically predicts the probabilities of several patient outcomes at different stages during hospital admission. 
	}
	\label{fig:illustration}
\end{figure}

Existing work on clinical note representation learning and multitask clinical outcome prediction focuses on building neural architectures for feature learning~\cite{huang2019clinicalbert} or considers different prediction tasks under the joint learning framework~\cite{harutyunyan2019multitask}.
However, there remain some unsolved challenges. 
First, the joint learning scheme for multiple tasks cannot effectively deal with inter-task interference, in which the predictors of different tasks compete with each other. 
Second, the task context information is usually underused, making multitask learning task-ignorant. 
Third, current learning algorithms are easy to overfit on seen training examples but fail to generalize on rare diseases or unseen diagnoses. 

We propose a novel multitask learning method for patient outcome and zero-shot diagnosis prediction from clinical notes to solve the aforementioned challenges. 
We propose to incorporate task information to enable task-specific prediction in multitask learning.
Inspired by the hypernetworks~\cite{ha2017hypernetworks} that use a small neural network to generate parameters for a larger neural network, we encode the semantic task information as the task context and use the encoded task embeddings as shared meta knowledge to generate the task-specific parameters of prediction heads for multiple tasks.
Our proposed method effectively utilizes the task information and produces task-aware multitask prediction heads. 
Our method is also generalizable on unseen diagnoses by maintaining the shared meta knowledge encoded as semantic task embeddings. 
Furthermore, we propose to dynamically update the weight coefficients of the multitask learning objective using another random hypernetwork, inspired by random projections~\cite{wojcik2019training}, to balance the learning from multiple clinical tasks.

Our contributions include: 
\begin{itemize}
	\item We propose to utilize task information via task embeddings for multitask patient outcome prediction spanning readmission, diagnosis, and length-of-stay prediction from clinical text and develop a hypernetwork for task-specific parameter generation to share information among different tasks. 
	\item We propose to regularize the objective function via a randomly parameterized task weighting scheme that effectively balances the learning process among multiple tasks, taking their relationships into account.
	\item Experiments on two datasets extracted from a real-world clinical database show that our proposed method outperforms strong multitask learning baselines and achieves more robust performance in zero-shot diagnosis prediction. 
\end{itemize}

\section{Method}

\subsection{Problem Setup and Overall Architecture}
\label{sec:problem_definition}

\begin{figure*}[htbp!]
\centering
	\includegraphics[width=0.9\textwidth]{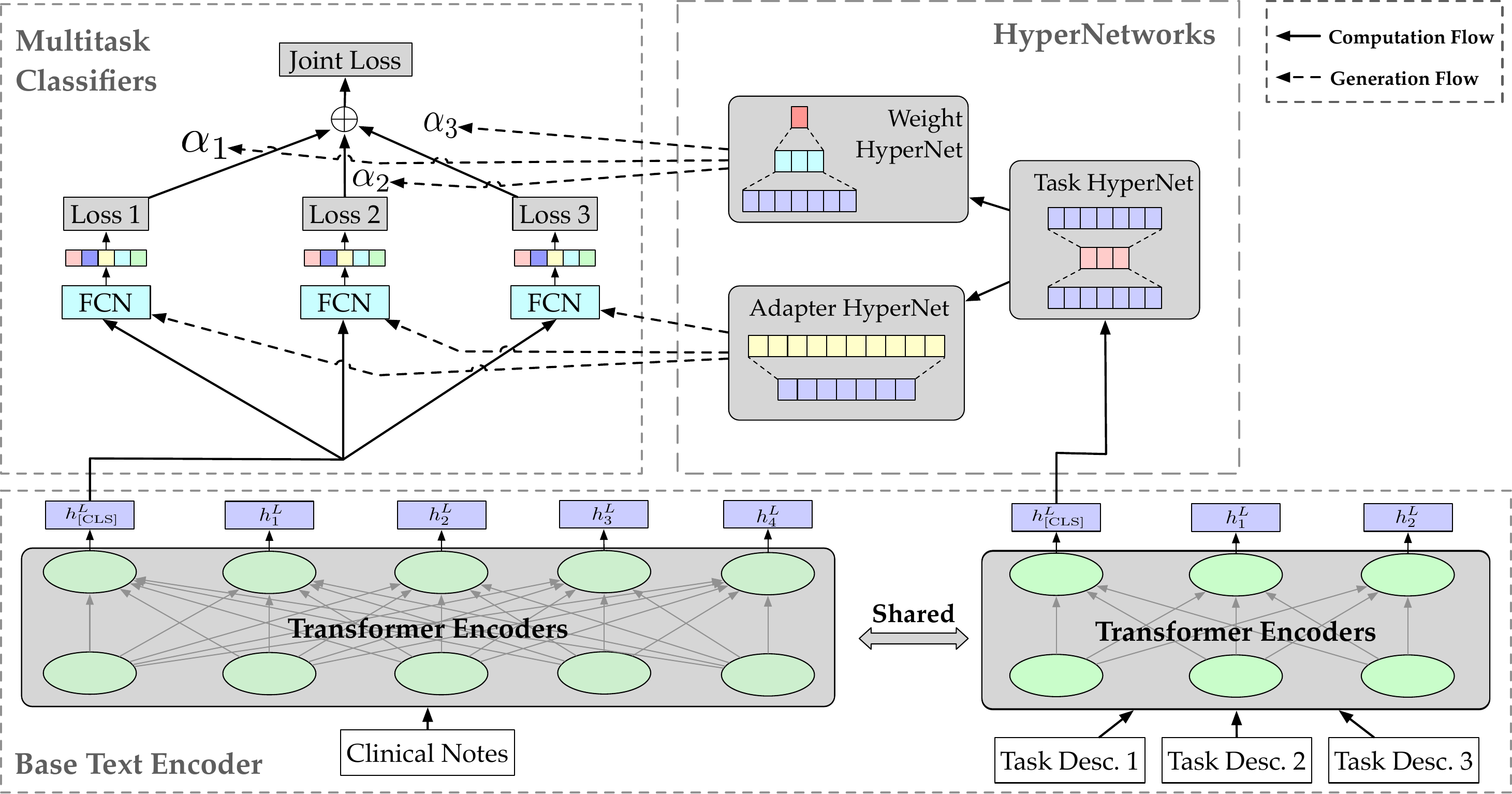}
	\caption{The illustration of our proposed hypernetwork-guided multitask learning framework}
	\label{fig:framework}
\end{figure*}

The clinical document in patient health records is denoted as $\mathbf{d}_{1:n}=x_{1}, \ldots, x_{n}$, where each $x_i$ is the $i$-th token.
Patient outcome prediction predicts clinical results by mapping the input text into prediction scores with a function $\mathcal{F}: \mathcal{X}^{n} \rightarrow \mathcal{Y}^m$ such that
$y=\mathcal{F}\left(x_1, \ldots, x_n; \boldsymbol{\theta}\right)$, where $y\in \mathbb{R}^m$ is the patient outcome, $m$ is the number of classes, and $\boldsymbol{\theta}$ are the model parameters.
Under the multitask task setting, we have $k$ learning tasks $\{ \mathcal{T}_t \}^{k}_{t=1}$ that are related to the outcome of clinical intervention. 
Multitask learning algorithm captures the relatedness of multiple tasks and improves the modeling~\cite{zhang2021survey}. 
In this study, we predict multiple clinical outcomes. 
Thus, the goal of the learning process is to fit a single-input multi-output function $y_t=\mathcal{F}\left(x_1, \ldots, x_n; \boldsymbol{\theta}_{shared}, \boldsymbol{\theta}_t \right)$, where $y_t\in \mathbb{R}^{m_t}$ is the ground truth label and $m_t$ is the number of classes of the $t$-th task, $\boldsymbol{\theta}_{shared}$ are the shared model parameters, and $\boldsymbol{\theta}_t$ the task-specific parameters.
The learning objective is to minimize the weighted sum of loss functions of multiple tasks, denoted as
\begin{equation}
	\mathcal{L}(\boldsymbol{\theta})=\sum_{t=1}^{k} \alpha_{t} \mathcal{L}_t\left(y_{t}, \hat{p}_{t} ; \boldsymbol{\theta}_{shared}, \boldsymbol{\theta}_t\right),
\end{equation}
where $\alpha_t$ is the weight coefficient of $t$-th task.  

The progress of a clinical intervention comes with multiple records, as illustrated in Fig. \ref{fig:illustration}, leading to lengthy clinical notes.
Following~\citet{huang2019clinicalbert}, we segment long clinical documents into several chunks, i.e., the document segmented to $r$ chunks becomes $\mathbf{d}_{1:r}=c_{1}, \ldots, c_i, \ldots, c_{r}$, where the $i$-th chunk $c_i$ contains $s$ tokens.
The segmentation has two advantages: 
1) it enables progressive prediction from chunks of clinical notes calculated from the predictions of individual chunks (see details in Experiments); 
2) it makes the input text length suitable for standard transformer-based pre-trained models (described in the next section).

We propose a hypernetwork-guided multitask learning framework to predict multiple patient outcomes. 
Fig. \ref{fig:framework} illustrates the overall framework, where the generation flow generates parameters. 
Specifically, we use a shared transformer-based text encoder to obtain hidden representations of clinical notes and task information. 
We develop a hypernetwork-based module, called Adapter HyperNet, to generate task-specific parameters for classification heads based on the task embeddings trained by a bottleneck network. 
The generated task-specific parameters then become the parameters of classification heads. 
To balance the learning of the joint objective function, we deploy the weight hypernetwork to generate weight coefficients conditioned on the task embeddings.

\subsection{Base Text Encoder}
\label{sec:base_model}

We deploy the Bidirectional Encoder Representations from Transformers (BERT) model~\cite{devlin2019bert} as our base text encoder to learn rich text features. 
The BERT text encoder pre-trains a language model with masked language modeling and next sentence prediction objectives in an unsupervised manner.
Then, a downstream fine-tuning is followed by tuning the pre-trained model checkpoint with suitable learning objectives for downstream applications.
The pre-training and fine-tuning paradigm can effectively exploit semantic knowledge from large training corpora and achieve superior performance in many downstream applications.
Due to the discrepancy of vocabularies in general and specific domains, many efforts have been made to pre-train a transformer language model in various specific domains.
To benefit the most from unsupervised representation learning for clinical application, we adopt the ClinicalBERT that starts from the standard BERT checkpoint and continues pre-training on clinical notes~\cite{huang2019clinicalbert}.
Given text chunks $c_i$ as inputs, we use the last layer's hidden representation of BERT-based text encoder to represent the encoded clinical note, i.e.,
\begin{equation}
	\mathbf{H} = \left[\operatorname{BERT}(c_1,~\boldsymbol{\theta_{0}}),~\dots, \operatorname{BERT}(c_r,~\boldsymbol{\theta_{0}})\right]
\end{equation}
where $\mathbf{H} \in \mathbb{R}^{s\times d_h}$, $s$ is the sequence length, $c_i$ is the $i$-th chunked text, and $\boldsymbol{\theta_0}$ is the model parameter of BERT encoder initialized from self-supervised pretraining. 
We can use the pooled embedding of the hidden representations to represent the clinical note, which is denoted as $\mathbf{h}\in \mathbb{R}^{d_h}$.
Similarly, we input the textual label of each task to obtain the task description embeddings with a shared BERT-based text encoder parameterized with $\boldsymbol{\theta_{0}}$.
Specifically, for the $t$-th task with $m_t$ classes, the initial task embeddings are $\mathbf{T_t}\in \mathbb{R}^{m_t \times d_t}$, where $d_t$ is the embedding dimension.
The obtained task embeddings capture the task information and preserve the semantic meaning of classes in each task. 

\subsection{Task-Conditioned Hypernetworks}

Our goal is to utilize task side-information to enable robust patient outcome prediction, especially for those classes with very few instances or that were unseen in the training set. 
In the joint training of multitask learning, it is easy to suffer from intertask interference. 
Inspired by the hypernetworks~\cite{ha2017hypernetworks} that generate model parameters to enable weight-sharing across neural layers, we propose the task-conditioned hypernetworks with semantic task label embeddings to generate the parameters of multitask classification heads for clinical outcomes. 
The task-conditioned parameter generation can enable task context-aware learning from multiple tasks and share knowledge across different tasks.

\paragraph{Task Embeddings}
To facilitate the parameter generation conditioned on task information, we further use a bottleneck network to train the task embeddings obtained from the contextualized embeddings of the BERT-based text encoder.
The bottleneck network contains two fully-connected layers with ReLU as the activation function~\cite{nair2010rectified}, denoted as:
\begin{equation}
	\mathbf{Z_t} = \big(\operatorname{ReLU}(\mathbf{T_t}\mathbf{W_1})\big)\mathbf{W_2},
\end{equation}
where $\mathbf{W_1} \in \mathbb{R}^{d_t\times d_b}$ and $\mathbf{W_2} \in \mathbb{R}^{d_b\times d_t}$ are weight matrices of linear layers, $\mathbf{Z_t}\in \mathbb{R}^{m_t \times d_t}$ are the output task embeddings, $d_b$ is the hidden dimension that restricts the bottleneck with fewer neurons, and the bias terms are omitted for simplicity.   
The bottleneck network has been applied in many fields, such as image noise reduction. 
We use this structure to reduce the potential noise in task classes because the assignment of diagnosis in MIMIC-III does not use a systematic ontology.
 
\paragraph{Adapter Hypernetwork}
Inspired by the hypernetwork-generated adapters~\cite{mahabadi2021parameter} that fine-tune transformer layers, we equip the multitask classification heads with adapter hypernetworks that generate the parameters.
Furthermore, we inject the contextual information in the task labels into parameter generation, which provides task-specific classification heads capable of task context-aware learning.

The adapter hypernetwork takes task embeddings of each task as input and generates model parameters written as
\begin{equation}
	\mathbf{W_t} = \mathcal{H}(\mathbf{Z_t}),~\forall~t=1, \ldots, k
\end{equation}
where $\mathbf{W_t}\in \mathbb{R}^{d_h \times m_t}$ is the generated weight parameters for the classification head of the $t$-th task. 
We instantiate hypernetworks with simple linear layers. 
Specifically, the matrices of task embeddings $\mathbf{Z_t}$ are flattened into vectors $\mathbf{v_t} \in \mathbb{R}^{m_t * d_t \times 1}$, projected into the hidden representation space $\mathbb{R}^{m_t *d_h \times 1}$ with nonlinear activation, and reshaped to the size of weight matrices $\mathbb{R}^{m_t  \times d_h}$ to act as the parameters of classification heads.
The bias parameters are also generated similarly. 
The adapter hypernetwork conditioned on the task context shares the meta knowledge of tasks and enables task-specific prediction. 
Thus, the inter-task interference in the multitask setting can be effectively mitigated. 
Furthermore, the semantic embeddings of task context provide the adapter hypernetwork a good initialization of task classes to generate task-specific parameters, facilitating the meta-learning from limited information for zero- or few-shot scenarios. 

\subsection{Task-Conditioned Learning Objective}
\label{sec:objective}
In the joint training framework of multitask learning, the standard method corresponds to manually tuning the weight coefficient $\alpha_t$. 
However, this requires human efforts to configure additional hyperparameters when tuning the multitask learning algorithms manually. 
To avoid this, we propose to use another \textit{random} hypernetwork $\mathcal{G}$ to generate the weight coefficients and make the weighted objective function of the joint multitask learning conditioned on task context.
The weight coefficient hypernetwork $\mathcal{G}$ is defined as
\begin{equation}
	\beta_t = \mathcal{G}(\mathbf{Z_t}),~\forall~t=1, \ldots, k.
\end{equation}
This hypernetwork is a multilayer perceptron (MLP), and it takes the task embeddings $\mathbf{Z_t}$ as input and outputs a one-dimensional scalar. 
Then we apply the softmax function to generate normalized weight probability, i.e., 
\begin{equation}
	\alpha_t = \frac{\exp (\beta_t)}{\sum_{t=1}^{k}\exp{\beta_t}}.
\end{equation}
In practice, we initialize the weight hypernetwork randomly and freeze the weights during optimization. 
Hence this approach can be seen as analogous to the technique of random projection~\cite{wojcik2019training}, which projects high-dimensional points to a lower dimension using a random weight matrix, and which is known to preserve the relationships of the points. 
Here, however, the role of the projection weight matrix is taken by the weight hypernetwork. 
In this way, the weight coefficients learned from the task context can preserve the semantic task information and reweigh the joint loss dynamically when the contexts change. 
We hypothesize that this also adds flexibility to training and regularization of the model, which contributes to the improved performance (see Section \ref{sec:ablation}).

\section{Experiments}

\subsection{Datasets and Setup}

\subsubsection{Datasets}
We use the admission-level patient records in the MIMIC-III dataset for experiments. 
From the vast information in this database, we select the free-text patient notes as the testbed for natural language processing research. 
We release the source code \footnote{Code available at \url{https://agit.ai/jsx/MT-Hyper}}.

\begin{table}[htbp!]
\centering
\scriptsize
\setlength{\tabcolsep}{2pt}
\caption{A statistical summary of datasets and tasks}
\label{tab:data}
\begin{tabular}{l | c c c |cccc }
\toprule
\multirow{2}{4em}{Dataset}		 & \multicolumn{3}{c}{\# Samples}  & \multicolumn{4}{c}{\# Classes} \\
\cline{2-8} 
& Train & Val. & Test & Readm. & Adm. Type & Diag. & LoS \\
\midrule
Discharge & 26,245 & 3,037 & 3,063 & 2 & 3 & 2,624 & 10 \\
Early Notes & 6,656 & 743 & 608 & 2 & 3 & 2,637 & 10 \\
\bottomrule
\end{tabular}
\end{table}

This study considers four classification tasks, i.e., readmission prediction, diagnosis classification, length-of-stay prediction, and admission type classification. 
Following the prior work~\cite{huang2019clinicalbert}, we build two datasets, i.e., one extracted from discharge summaries (denoted as \texttt{Discharge}) and one with early notes that were created within three days after admission (denoted as \texttt{Early Notes}), according to the period of patient admission date and the date when the notes are charted. 
Table \ref{tab:data} summarizes the number of instances and classes in our extracted datasets.  
In-hospital death prediction is also an essential task of patient outcome. 
However, we focus more on readmission, and mortality precludes the possibility of readmission. 
Thus, we filter out all the in-hospital death cases, which also aligns with the prior work~\cite{huang2019clinicalbert}.
We also consider a proxy task of admission type classification that categorizes the clinical notes into ``emergency", ``elective" and ``urgent" (studied in Sec.~\ref{sec:task_conditioning}).

\subsubsection{Tasks}
\paragraph{Readmission Prediction}
The goal of this binary classification task is to predict whether the patient will be readmitted within 30 days of discharge.

\paragraph{Diagnosis Prediction}
The MIMIC-III dataset has a total of 15,691 classes of diagnoses in the ``ADMISSIONS'' table.
We extract the clinical notes from the ``NOTEEVENTS" table and get a total of 2,715 diagnoses in the extracted datasets of discharge and early notes, where frequent diagnoses include pneumonia, congestive heart failure, and sepsis. 
In our train/val/test split of discharge notes, 2,209 diagnoses appear in the training set, which allows us to test the performance of zero-shot prediction. 

\paragraph{Length-of-Stay (LoS) Prediction}
Following the setting of \citet{harutyunyan2019multitask}, we define the length-of-stay prediction as a 10-class classification problem.
Specifically, the duration of stay is divided into ten buckets, i.e., one class for stays less than a day, seven classes for one-to-seven-day stays, one for stays longer than a week but shorter than two, and the last one class for stays over two weeks.
A slight difference is that the duration of stay in our definition is calculated during the span between when the patient was admitted to the hospital and when discharged from the hospital.

\begin{table*}[ht!]
\centering
\scriptsize
\setlength{\tabcolsep}{8pt}
\caption{Results of patient outcome prediction from early notes with average score $\pm$ standard deviation reported. Bold values indicate the cases when our model obtain the best performance.}
\label{tab:early}
\begin{tabular}{l | l | c c | cc }
\toprule
 \multirow{2}{4em}{Task} &  \multirow{2}{4em}{Method} & \multicolumn{2}{c}{Progressive} & \multicolumn{2}{c}{Ultimate} \\
 & & F1 & AUC-ROC & F1 & AUC-ROC \\
\midrule
\multirow{4}{6em}{Readmission}	&	MT-LSTM	& $	51.69	\pm	3.72	$&$	53.80	\pm	1.99	$&$	52.36	\pm	3.62	$&$	56.17	\pm	2.23	$ \\	
	&	MT-BERT	& $	55.52	\pm	3.92	$&$	56.75	\pm	2.56	$&$	56.02	\pm	4.34	$&$	60.85	\pm	1.61	$ \\	
	&	MT-RAM	& $	56.00	\pm	3.09	$&$	57.26	\pm	1.94	$&$	56.51	\pm	2.80	$&$	62.78	\pm	1.39	$ \\	
	&	MT-Hyper	& $	\textbf{56.38}	\pm	2.30	$&$	\textbf{57.57}	\pm	1.27	$&$	\textbf{57.67}	\pm	2.21	$&$	62.41	\pm	1.61	$ \\	\hline
\multirow{4}{6em}{Diagnosis}	&	MT-LSTM	& $	8.50	\pm	3.39	$&$	63.00	\pm	0.95	$&$	8.69	\pm	3.79	$&$	62.26	\pm	0.91	$ \\	
	&	MT-BERT	& $	11.63	\pm	4.83	$&$	64.13	\pm	1.30	$&$	11.80	\pm	5.02	$&$	63.24	\pm	1.30	$ \\	
	&	MT-RAM	& $	14.06	\pm	1.21	$&$	68.58	\pm	1.60	$&$	14.39	\pm	1.41	$&$	67.84	\pm	1.64	$ \\	
	&	MT-Hyper	& $	\textbf{19.56}	\pm	1.33	$&$	\textbf{72.98}	\pm	0.45	$&$	\textbf{20.47}	\pm	1.38	$&$	\textbf{73.21}	\pm	0.43	$ \\	\hline
\multirow{4}{6em}{LOS}	&	MT-LSTM	& $	30.19	\pm	2.70	$&$	75.73	\pm	0.92	$&$	30.54	\pm	2.68	$&$	76.50	\pm	0.87	$ \\	
	&	MT-BERT	& $	26.40	\pm	4.34	$&$	72.60	\pm	2.81	$&$	27.25	\pm	4.08	$&$	73.71	\pm	2.66	$ \\	
	&	MT-RAM	& $	26.20	\pm	3.08	$&$	73.15	\pm	3.44	$&$	27.08	\pm	2.80	$&$	74.04	\pm	3.31	$ \\	
	&	MT-Hyper	& $	\textbf{33.18}	\pm	0.91	$&$	71.84	\pm	1.95	$&$	\textbf{33.28}	\pm	1.03	$&$	72.23	\pm	2.18	$ \\	\hline
\multirow{4}{6em}{Average}	&	MT-LSTM	& $	30.12	\pm	3.27	$&$	64.18	\pm	1.29	$&$	30.53	\pm	3.36	$&$	64.97	\pm	1.34	$ \\	
	&	MT-BERT	& $	31.18	\pm	4.36	$&$	64.49	\pm	2.22	$&$	31.69	\pm	4.48	$&$	65.93	\pm	1.86	$ \\	
	&	MT-RAM	& $	32.09	\pm	2.46	$&$	66.33	\pm	2.33	$&$	32.66	\pm	2.34	$&$	68.22	\pm	2.11	$ \\	
	&	MT-Hyper	& $	\textbf{36.37}	\pm	1.51	$&$	\textbf{67.47}	\pm	1.23	$&$	\textbf{37.14}	\pm	1.54	$&$	\textbf{69.28}	\pm	1.41	$ \\	
\bottomrule
\end{tabular}
\end{table*}

\subsubsection{Baselines}
We compare our method, dubbed MT-Hyper, with several multitask learning algorithms on electronic health records and clinical natural language processing. 
\textbf{MT-LSTM~\cite{harutyunyan2019multitask}} adopts a basic LSTM-based neural network and is jointly trained with several patient outcome prediction tasks. 
This method is initially designed for modeling time series. 
We modify it with a word embedding module for modeling clinical text in patients' health records.
\textbf{MT-BERT~\cite{mulyar2020mt}} is a unified clinical NLP model that utilizes the BERT model architecture for text encoding and learns features with multiple clinical task prediction heads simultaneously. 
It originally consists of eight task heads, while our implementation uses fewer clinical prediction tasks.
\textbf{MT-RAM~\cite{sun2021multitask}} proposes a neural module with feature recalibration and aggregation for multitask learning that mitigates the mutual interference of different tasks and refines the feature learning from noisy clinical text.

\subsubsection{Settings and Evaluation}
We use the Adam optimizer~\cite{kingma2014adam} to optimize the model and set different learning rates for different text encoders and classification heads.
The learning rates of LSTM-based methods range from $5\mathrm{e}^{-5}$ to $1\mathrm{e}^{-3}$.
As to BERT-based methods, we use a lower learning rate for both base text encoder and classification heads, ranging from $1\mathrm{e}^{-5}$ to $5\mathrm{e}^{-5}$. 
Because clinical BERT only has pretrained checkpoints with the BERT base model and the limitation of computing resources, we adopt the BERT base architecture as our text encoder.
We use 300D word embedding, and the dimension of LSTM hidden states is also set to 300.
We run each algorithm ten times and report the average score and standard deviation. 
We monitor the validation loss in each trial and set the number of epochs that triggers the early stop mechanism to be 3 for the BERT encoder and 10 for the LSTM encoder.

Our problem definition considers readmission as a binary classification problem, while the rest of the tasks are defined as multi-class classification problems.
For readmission prediction, we use weighted F1 and AUC-ROC. 
For multi-class classification problems, we report micro averaged scores.
As we segment lengthy clinical notes into shorter chunks for progressive prediction, we report the results of two types of evaluation scheme, i.e., \textit{progressive} scores that are calculated with the prediction probabilities of note chunks and the \textit{ultimate} scores of a complete admission that are measured by aggregating every prediction on segmented notes.
The ultimate scores can also be viewed as a similar strategy to the bagging-based ensemble, in which each single prediction is based on different chunks of an instance and the final prediction is an ensemble of different predictions.

\subsection{Main Results}
This section presents the main results on prediction at discharge and early prediction.
As admission type classification has limited usage in real-world scenarios, we do not include it in the comparison of main results but use it in the discussion of task conditioning. 

\paragraph{Early Prediction}
Table~\ref{tab:early} shows the results of patient outcome prediction from early notes.
In both progressive and ultimate prediction, our proposed method outperforms three baselines in most cases.
The baseline models are relatively stronger in tasks with fewer classes.
For example, MT-BERT and MT-RAM produce a relatively good prediction on readmission, while their performance for multi-class diagnosis prediction is much worse than our proposed method.
Furthermore, our proposed method can make more stable predictions in most cases according to the standard deviation. 
Our method outperforms the baselines by a relatively large margin in terms of the average scores of three prediction tasks.

\paragraph{Prediction at Discharge}
Next, we consider prediction at discharge, while keeping the setup otherwise similar to the early prediction. 
Discharge notes provide relatively complete information about patient hospital visits.
Table~\ref{tab:discharge} shows the prediction results. 
Our method has comparable readmission prediction performance, superior diagnosis prediction, inferior performance in the length-of-stay prediction, and relatively better overall performance measured by average score than other baselines. 
Besides, our method has smaller values of standard deviation in most cases when making a prediction at discharge.

\begin{table*}[htbp]
\centering
\scriptsize
\setlength{\tabcolsep}{8pt}
\caption{Results of patient outcome prediction at discharge with average score $\pm$ standard deviation reported. Bold values indicate the cases when our model obtain the best performance.}
\label{tab:discharge}
\begin{tabular}{l | l | c c | cc }
\toprule
 \multirow{2}{4em}{Task} &  \multirow{2}{4em}{Method} & \multicolumn{2}{c}{Progressive} & \multicolumn{2}{c}{Ultimate} \\
 & & F1 & AUC-ROC & F1 & AUC-ROC \\
\midrule
\multirow{4}{6em}{Readmission}	&	MT-LSTM	& $	60.54	\pm	1.38	$&$	60.39	\pm	1.22	$&$	63.17	\pm	1.64	$&$	68.46	\pm	1.93	$ \\	
	&	MT-BERT	& $	64.86	\pm	1.08	$&$	64.78	\pm	1.15	$&$	68.66	\pm	3.17	$&$	76.33	\pm	2.88	$ \\	
	&	MT-RAM	& $	65.92	\pm	1.86	$&$	65.88	\pm	1.34	$&$	70.41	\pm	2.67	$&$	77.49	\pm	2.30	$ \\	
	&	MT-Hyper	& $	\textbf{66.41}	\pm	0.77	$&$	\textbf{66.08}	\pm	0.74	$&$	69.66	\pm	1.52	$&$	76.93	\pm	1.11	$ \\	\hline
\multirow{4}{6em}{Diagnosis}	&	MT-LSTM	& $	9.08	\pm	1.60	$&$	66.51	\pm	0.98	$&$	11.04	\pm	2.74	$&$	64.85	\pm	1.01	$ \\	
	&	MT-BERT	& $	9.75	\pm	1.82	$&$	69.19	\pm	6.22	$&$	10.80	\pm	2.83	$&$	68.04	\pm	7.98	$ \\	
	&	MT-RAM	& $	10.26	\pm	1.99	$&$	67.38	\pm	0.79	$&$	11.39	\pm	2.99	$&$	65.70	\pm	0.92	$ \\	
	&	MT-Hyper	& $	\textbf{10.28}	\pm	0.25	$&$	\textbf{76.93}	\pm	0.73	$&$	\textbf{12.99}	\pm	1.57	$&$	\textbf{79.07}	\pm	1.07	$ \\	\hline
\multirow{4}{6em}{LOS}	&	MT-LSTM	& $	31.09	\pm	0.93	$&$	76.77	\pm	0.67	$&$	32.57	\pm	1.29	$&$	79.33	\pm	1.19	$ \\	
	&	MT-BERT	& $	31.37	\pm	1.23	$&$	78.27	\pm	0.90	$&$	34.91	\pm	0.31	$&$	82.13	\pm	0.26	$ \\	
	&	MT-RAM	& $	31.29	\pm	1.91	$&$	76.14	\pm	4.48	$&$	32.29	\pm	2.95	$&$	77.69	\pm	5.34	$ \\	
	&	MT-Hyper	& $	31.33	\pm	0.38	$&$	73.76	\pm	1.31	$&$	31.08	\pm	0.21	$&$	73.01	\pm	1.28	$ \\	\hline
\multirow{4}{6em}{Average}	&	MT-LSTM	& $	33.57	\pm	1.30	$&$	67.89	\pm	0.96	$&$	35.59	\pm	1.89	$&$	70.88	\pm	1.38	$ \\	
	&	MT-BERT	& $	35.33	\pm	1.37	$&$	70.75	\pm	2.76	$&$	38.13	\pm	2.10	$&$	75.50	\pm	3.71	$ \\	
	&	MT-RAM	& $	35.82	\pm	1.92	$&$	69.80	\pm	2.20	$&$	38.03	\pm	2.87	$&$	73.63	\pm	2.86	$ \\	
	&	MT-Hyper	& $	\textbf{36.01}	\pm	0.47	$&$	\textbf{72.26}	\pm	0.93	$&$	37.91	\pm	1.10	$&$	\textbf{76.34}	\pm	1.15	$ \\	
\bottomrule
\end{tabular}
\end{table*}

\subsection{Zero-shot Diagnosis Prediction}
We study a more challenging scenario of zero-shot diagnosis prediction.
In the extracted datasets, among the total 2,624 diagnoses of discharge, 2,209 diagnoses appear in the training set, i.e., 415 types of diagnoses are unseen in the evaluation data. 
We adopt this sampled setting to test the performance of zero-shot diagnosis prediction.
Table~\ref{tab:zero-shot-diagnosis} shows the AUC-ROC scores of progressive and ultimate diagnosis prediction on zero-shot diagnosis categories. 
The results indicate that the baseline models make terribly incorrect predictions with AUC-ROC scores lower than the random guess. 
These baselines cannot learn meta-knowledge of diagnoses, making them overfitting on seen diagnoses in the training set. 
In contrast, our proposed MT-Hyper can capture the task context and semantic diagnosis information. 
Therefore, our method is more generalizable on unseen diagnoses and achieves satisfactory performance under the challenging zero-shot setting.

\begin{table}[htbp!]
\centering
\scriptsize
\caption{Results (AUC-ROC) of zero-shot diagnosis prediction on unseen diagnosis results with average score $\pm$ standard deviation reported. Bold values indicate the cases when our model obtain the best performance.}
\setlength{\tabcolsep}{8pt}
\label{tab:zero-shot-diagnosis}
\begin{tabular}{l | l | c c }
\toprule
Dataset &  Method & Progressive & Ultimate \\
\midrule
\multirow{4}{5em}{Discharge}	&	MT-LSTM	& $					11.00	\pm	2.31	$ & $					9.74	\pm	1.55	$ \\
	&	MT-BERT	& $					10.47	\pm	0.76	$ & $					9.00	\pm	0.36	$ \\
	&	MT-RAM	& $					11.40	\pm	1.44	$ & $					10.37	\pm	1.25	$ \\
	&	MT-Hyper	& $					\textbf{64.06}	\pm	2.02	$ & $					\textbf{68.33}	\pm	2.76	$ \\
\hline
\multirow{4}{5em}{Early}	&	MT-LSTM	& $					8.65	\pm	1.19	$ & $					8.47	\pm	1.00	$ \\
	&	MT-BERT	& $					8.24	\pm	0.79	$ & $					8.20	\pm	0.74	$ \\
	&	MT-RAM	& $					14.25	\pm	3.11	$ & $					14.12	\pm	3.12	$ \\
	&	MT-Hyper	& $					\textbf{62.04}	\pm	1.13	$ & $					\textbf{63.76}	\pm	1.03	$ \\

\bottomrule
\end{tabular}
\end{table}

\subsection{Semantic Relation of Task Embeddings}
\label{sec:task_conditioning}

\begin{figure}[htbp]
\begin{center}
  \begin{subfigure}[]{0.25\textwidth}
  	\includegraphics[width=\textwidth]{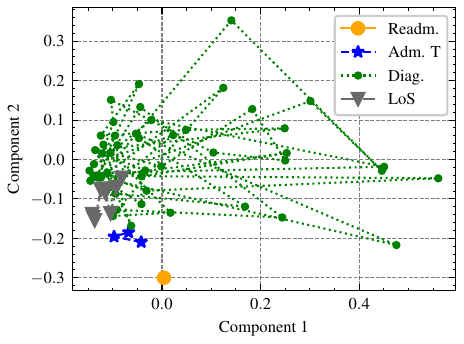} 
	\subcaption{Task embedding}
	\label{fig:task_embedding_all}
  \end{subfigure}%
  \begin{subfigure}[]{0.25\textwidth}
  	\includegraphics[width=\textwidth]{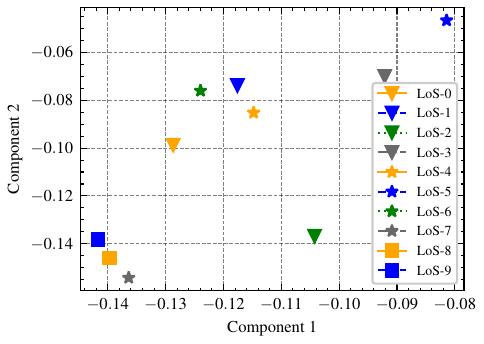}
	\subcaption{LOS label embeddings}
	\label{fig:task_embedding_los}
  \end{subfigure}
\caption{Visualization of task embeddings with dimension reduced by PCA}
\label{fig:task_embed}
\end{center}
\end{figure}

We study the semantic relation of task embeddings to verify if task conditioning uses task information. 
We include the admission type classification as an auxiliary task and train four tasks jointly to achieve this goal. 
We filter out ``newborn" and ``death" and define the admission type classification as a three-way classification task with admission types of ``emergency", ``elective" and ``urgent". 
The classification task categorizes the type of admission for each clinical note. 
It may have no practical usage in clinical support.
However, it is a useful auxiliary task for clinical note understanding. 
Thus, we choose this task as an auxiliary task to help study the semantic relation of task embeddings.
We obtain the task embeddings from the trained model and aggregate embeddings of each task into a vector via mean pooling. 
To visualize task vectors in low dimensional space, we apply principal component analysis to reduce the dimension of task vectors. 
Firstly, we plot task label embeddings of readmission, admission type and length-of-stay and together with the first 50 diagnoses. 
Figure \ref{fig:task_embedding_all} shows that different label embeddings are relatively far apart from each other, and the two tasks that are similar to each other, readmission prediction and admission type prediction, appear close to each other. 
We further zoom in to the embeddings of length-of-stay task labels in Figure \ref{fig:task_embedding_los}.
However, here we cannot find any clear interpretable pattern. 
We hypothesize that the prediction of length-of-stay would benefit from clinical document understanding and numerical reasoning over the ordinal categories, and the hypernetwork-based architecture with semantic task embeddings does not effectively capture this ordinal relation of different lengths of stay. 
This may partially explain why our method did not provide advantage over other methods on the length-of-stay prediction.

\subsection{Ablation Study}\label{sec:ablation}
\begin{table*}[ht!]
\centering
\scriptsize
\setlength{\tabcolsep}{8pt}
\caption{Ablation study on patient outcome prediction from early notes with three tasks jointly trained}
\label{tab:ablation_early}
\begin{tabular}{l | l | c c | cc }
\toprule
 \multirow{2}{4em}{Task} &  \multirow{2}{4em}{Method} & \multicolumn{2}{c}{Progressive} & \multicolumn{2}{c}{Ultimate} \\
 & & F1 & AUC-ROC & F1 & AUC-ROC \\
\midrule
\multirow{2}{6em}{Readmission}	&	without WH	& $	56.74	\pm	1.26	$&$	57.28	\pm	1.17	$&$	57.94	\pm	1.53	$&$	61.56	\pm	1.79	$ \\	
	&	with WH	& $	56.38	\pm	2.30	$&$	\textbf{57.57}	\pm	1.27	$&$	57.67	\pm	2.21	$&$	\textbf{62.41}	\pm	1.61	$ \\	\hline
\multirow{2}{6em}{Diagnosis}	&	without WH	& $	18.90	\pm	2.06	$&$	69.26	\pm	0.33	$&$	19.91	\pm	2.29	$&$	69.58	\pm	0.47	$ \\	
	&	with WH	& $	\textbf{19.56}	\pm	1.33	$&$	\textbf{72.98}	\pm	0.45	$&$	\textbf{20.47}	\pm	1.38	$&$	\textbf{73.21}	\pm	0.43	$ \\	\hline
\multirow{2}{6em}{LOS}	&	without WH	& $	33.59	\pm	0.36	$&$	71.14	\pm	1.25	$&$	33.85	\pm	0.34	$&$	71.44	\pm	1.32	$ \\	
	&	with WH	& $	33.18	\pm	0.91	$&$	\textbf{71.84}	\pm	1.95	$&$	33.28	\pm	1.03	$&$	\textbf{72.23}	\pm	2.18	$ \\	\hline
\multirow{2}{6em}{Average}	&	without WH	& $	36.41	\pm	1.23	$&$	65.89	\pm	0.92	$&$	37.23	\pm	1.39	$&$	67.53	\pm	1.19	$ \\	
	&	with WH	& $	36.37	\pm	1.51	$&$	\textbf{67.46}	\pm	1.22	$&$	37.14	\pm	1.54	$&$	\textbf{69.28}	\pm	1.41	$ \\	
\bottomrule
\end{tabular}
\end{table*}

\begin{table*}[ht!]
\centering
\scriptsize
\setlength{\tabcolsep}{8pt}
\caption{Ablation study on patient outcome prediction at discharge with three tasks jointly trained}
\label{tab:ablation_discharge}
\begin{tabular}{l | l | c c | cc }
\toprule
 \multirow{2}{4em}{Task} &  \multirow{2}{4em}{Method} & \multicolumn{2}{c}{Progressive} & \multicolumn{2}{c}{Ultimate} \\
 & & F1 & AUC-ROC & F1 & AUC-ROC \\
\midrule
\multirow{2}{6em}{Readmission}	&	without WH	& $	65.42	\pm	0.88	$&$	65.21	\pm	0.73	$&$	68.11	\pm	1.46	$&$	75.99	\pm	0.37	$ \\	
	&	with WH	& $	\textbf{66.41}	\pm	0.77	$&$	\textbf{66.08}	\pm	0.74	$&$	\textbf{69.66}	\pm	1.52	$&$	\textbf{76.93}	\pm	1.11	$ \\	\hline
\multirow{2}{6em}{Diagnosis}	&	without WH	& $	10.05	\pm	0.87	$&$	75.13	\pm	1.12	$&$	12.68	\pm	1.39	$&$	78.54	\pm	1.56	$ \\	
	&	with WH	& $	\textbf{10.28}	\pm	0.25	$&$	\textbf{76.93}	\pm	0.73	$&$	\textbf{12.99}	\pm	1.57	$&$	\textbf{79.07}	\pm	1.07	$ \\	\hline
\multirow{2}{6em}{LOS}	&	without WH	& $	31.20	\pm	0.31	$&$	73.47	\pm	0.59	$&$	30.82	\pm	0.34	$&$	72.62	\pm	0.79	$ \\	
	&	with WH	& $	\textbf{31.33}	\pm	0.38	$&$	\textbf{73.76}	\pm	1.31	$&$	\textbf{31.08}	\pm	0.21	$&$	\textbf{73.01}	\pm	1.28	$ \\	\hline
\multirow{2}{6em}{Average}	&	without WH	& $	35.56	\pm	0.69	$&$	71.27	\pm	0.81	$&$	37.20	\pm	1.06	$&$	75.72	\pm	0.91	$ \\	
	&	with WH	& $	\textbf{36.01}	\pm	0.47	$&$	\textbf{72.26}	\pm	0.93	$&$	\textbf{37.91}	\pm	1.10	$&$	\textbf{76.34}	\pm	1.15	$ \\	
\bottomrule
\end{tabular}
\end{table*}

Our proposed method consists of two hypernetworks, i.e., adapter hypernetwork for task-specific parameter generation and weight hypernetwork (WH) for task weight coefficient generation. 
We conduct an ablation study on the usage of weight hypernetwork. 
Table~\ref{tab:ablation_early} and Table~\ref{tab:ablation_discharge} show the results of patient outcome prediction from early notes and prediction at discharge respectively. 
The weight hypernetwork dynamically adjusts the weight coefficients of joint loss and contributes to the performance improvement in most cases.
MT-Hyper with weight hypernetwork has better predictive scores in 10 out of 16 evaluation metrics on patient outcome prediction from early notes.
As for patient outcome prediction at discharge, MT-Hyper with weight hypernetwork outperforms its counterpart without weight hypernetwork.

\section{Related~Work}

Patient outcome prediction, including automatic coding~\cite{friedman2004automated,yan2010medical}, patient severity~\cite{naemi2020prediction}, in-hospital mortality, decompensation, length of stay and phenotyping~\cite{harutyunyan2019multitask}, can empower health practitioners to make better clinical decision.
Many studies have been conducted to investigate the usability of deep learning for accurate prediction using time-series, clinical text, or multimodal data.
\citet{zufferey2015performance} compared different multi-label learning algorithms.
\citet{che2015deep} proposed a deep phenotyping model regularized on the categorical structure of the medical ontology.
\citet{zhang2019attain} proposed an attention-based LSTM network to model the disease progression in a time-aware fashion.
Similarly, \citet{men2021multi} extended the LSTM model to be attention- and time-aware for multi-disease prediction.
Convolutional neural networks have also been deployed in this field. 
For example, \cite{bardak2021improving} developed CNN-based networks for multimodal learning from patient records. 
Clinical notes as an essential modality of EHRs have also been studied intensively. 
\citet{ghassemi2014unfolding} extracted topic features from free-text patient data for mortality prediction.
\citet{mulyar2020mt} extracted clinical information from clinical documents via fine-tuning large-scale pretrained language models.

Multitask learning learns multiple tasks simultaneously with a shared representation and transfers the domain knowledge in related tasks to improve the learning capacity~\cite{caruana1997multitask}. 
It has also attracted much attention from the research community of machine learning and natural language processing for healthcare. 
\citet{mahajan2020identification} studied semantic similarity degrees of clinical notes by a multitask learning method with iterative data selection. 
\citet{sun2021multitask} proposed a multitask learning framework with feature calibration for medical coding. 
Several multitask learning schemes have also been developed for learning from clinical time-series~\cite{harutyunyan2019multitask}.
Hypernetworks~\cite{ha2017hypernetworks} that generate weights for other networks have also been adopted to solve the learning problem with multiple tasks, for example, the Hyperformer that generates the parameters of adapter modules for fine-tuning NLP task streams~\cite{mahabadi2021parameter}.
Inspired by this work, our work starts from a different setting that learns different tasks simultaneously, utilizes the semantic information of clinical predication tasks, and balances the learning objective via task conditioning.

\section{Conclusion}

Patient hospitalization is a complex process, which needs the learning algorithm to model multiple risk indicators. 
Multitask learning can jointly learn from patient records to empower clinical decision-making. 
To address the challenges of inter-task interference and poor generalizability to unseen labels in recent multitask learning frameworks for healthcare, we propose a hypernetwork-guided multitask learning method that learns from the task context and generates task-specific parameters for effective multitask prediction on patient outcomes. 
The proposed method incorporates contextualized language representations to encode clinical notes and capture the task context via the semantic embeddings of tasks.
The hypernetwork-guided multitask learning framework enables task-conditioned learning and balances the joint learning objective across different tasks.
Experimental studies on the real-world MIMIC-III dataset show our proposed method achieves better performance on early prediction and more robust performance on zero-shot diagnosis prediction.

\section*{Limitations}
The proposed task-conditioned multitask learning method requires to obtain task embeddings from the semantic description.

\section*{Acknowledgments}

We acknowledge the computational resources provided by the Aalto Science-IT project and CSC - IT Center for Science, Finland.
This work was supported by the Academy of Finland (Flagship programme: Finnish Center for Artificial Intelligence FCAI, and grants 336033, 352986) and EU (H2020 grant 101016775 and NextGenerationEU).

\bibliography{references.bib}
\bibliographystyle{acl_natbib}

\end{document}